\documentclass{article}

\usepackage[final]{corl_2019} 
\usepackage[utf8]{inputenc}
\usepackage{amssymb}
\usepackage{hyperref}
\usepackage{enumerate}
\usepackage{graphicx}
\usepackage{listings}
\usepackage{caption}
\usepackage{subcaption}
\usepackage{algorithm}
\usepackage{algorithmic}
\usepackage{booktabs}
\usepackage{multirow}
\usepackage{array}
\usepackage{pifont}
\usepackage{microtype}
\usepackage{xifthen}
\usepackage{textcomp}

\newcommand{\mb}{H_{\emptyset}}
\newcommand{\mi}{H_{i}}
\newcommand{\mih}{H_{ih}}
\newcommand{\mihg}{H_{ihg}}
\newcommand{\tfsz}{\footnotesize}

\newcommand{\ig}{{\texttt{gs}}}
\newcommand{\ir}{{\texttt{rgb}}}
\newcommand{\ip}{{\texttt{ps}}}
\newcommand{\ls}{\texttt{single}}
\newcommand{\ld}{\texttt{double}}
\newcommand{\lo}{\texttt{ordinal}}

\hypersetup{
    colorlinks=true,
    linkcolor=blue,
    filecolor=magenta,      
    urlcolor=cyan,
}

\title{Conditional Driving from \\Natural Language Instructions}
\author{
  Junha Roh$^{\dagger}$, Chris Paxton$^{\ddagger}$, Andrzej Pronobis$^{\dagger*}$, Ali Farhadi$^{\dagger\S\textsection}$, Dieter Fox$^{\dagger\ddagger}$\\
  University of Washington$^{\dagger}$, NVIDIA$^{\ddagger}$\\KTH Royal Institute of Technology$^{*}$, Allen Institute for AI$^{\S\textsection}$\\
  \texttt{\{rohjunha,pronobis,ali\}@cs.washington.edu}\\
  \texttt{\{cpaxton,dieterf\}@nvidia.com}
}

\begin{document}
\maketitle


\begin{abstract}
Widespread adoption of self-driving cars will depend not only on their safety but largely on their ability to interact with human users. Just like human drivers, self-driving cars will be expected to understand and safely follow natural-language directions that suddenly alter the pre-planned route according to user's preference or in presence of ambiguities, particularly in locations with poor or outdated map coverage. To this end, we propose a language-grounded driving agent implementing a hierarchical policy using recurrent layers and gated attention. The hierarchical approach enables us to reason both in terms of high-level language instructions describing long time horizons and low-level, complex, continuous state/action spaces required for real-time control of a self-driving car. We train our policy with conditional imitation learning from realistic language data collected from human drivers and navigators. Through quantitative and interactive experiments within the CARLA framework, we show that our model can successfully interpret language instructions and follow them safely, even when generalizing to previously unseen environments. Code and video are available at: \url{https://sites.google.com/view/language-grounded-driving}.
\end{abstract}


\begin{figure}[ht]
\centering
\includegraphics[width=0.8\columnwidth]{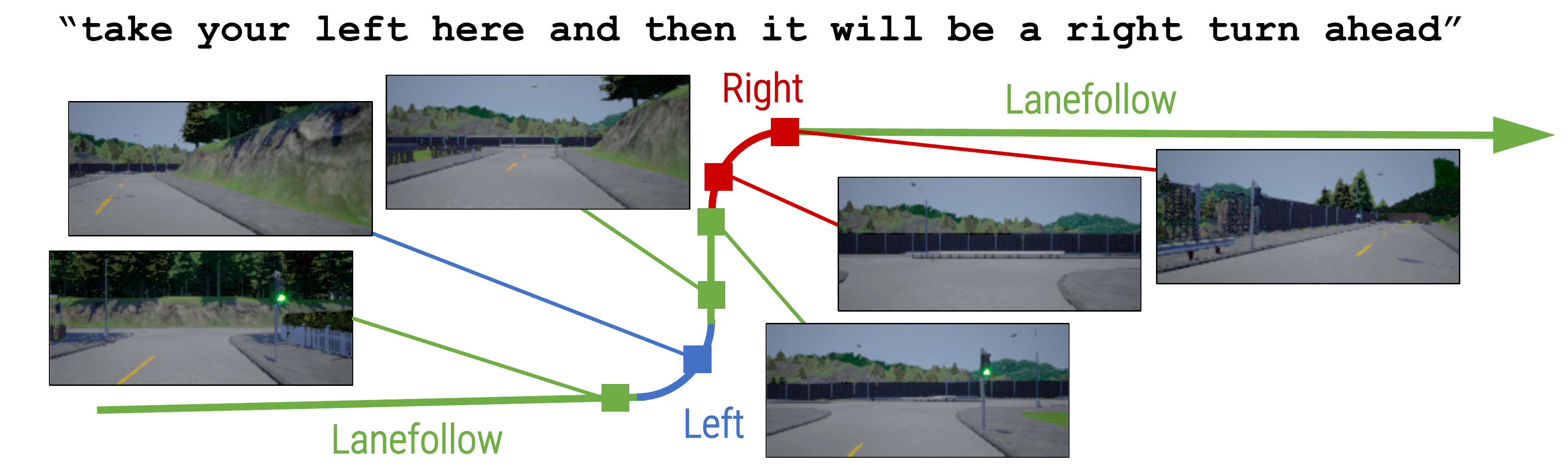}
\caption{Natural language control of self-driving vehicles. The user provides a high-level instruction; the vehicle must then (a) translate natural language into the correct sequence of high-level sub-tasks and (b) correctly execute these, by steering and applying the throttle as appropriate.}
\label{fig:cover}
\end{figure}

\section{Introduction}
\label{sec:introduction}
Passengers of self-driving cars will expect to interact with their vehicles in the same way as they do with human ride-share drivers. This includes providing specific directions to the precise drop-off locations, preferences about the chosen route, or clarifications in case of ambiguities. Furthermore, a car equipped with a skill to interpret natural-language, able to rely on the help of its user, will be more robust to navigation errors resulting from poor map coverage and inaccurate information about dynamic road conditions.

As shown in Fig.~\ref{fig:cover}, our goal is to learn to understand language instructions, such as ``you are going to go a little bit further for one block and make a left at the intersection,'' and use them to condition a policy that will drive a car safely using only image observations. The problem of end-to-end policy learning for self-driving cars is often formulated as imitation learning~\cite{bojarski2016end,xu2017end,codevilla2018end,muller2018driving}. We take a hierarchical approach. We use conditional imitation learning to learn policies for steering and throttle control, as in prior work~\cite{codevilla2018end,muller2018driving}. However, we also learn a high-level policy, which predicts high-level actions based on language instructions. This leads to a solution able to translate language into actions executed over long time horizons, including navigating multiple streets and turns before reaching the destination.

At the same time, we need to ensure that safety is not compromised, even in presence of incorrect and misleading instructions. A self-driving car will be used by non-experts, who might instruct the car to execute maneuvers that are not safe given the state of the world (e.g. to turn left when no left turn is possible). This is a known problem in language-to-control~\cite{paxton2019prospection}. Moreover, artificial systems often struggle with understanding the intricacies of realistic human language, in particular for such a dynamic task as driving. We provide two pathways to mitigate the harm this can cause: first, our policy is designed to ensure that only safe actions are taken, even when invalid input is given by the passenger. Second, the agent will complete maneuvers, such as driving through an intersection, even if new instructions from the user interrupt the current high-level plan.

We validate our proposed approach using CARLA~\cite{Dosovitskiy17}, an open-source driving simulator. We perform a set of quantitative transfer experiments, showing that our hierarchical models navigate correctly and safely, and can generalize between different environments. Furthermore, we perform a series of ablation tests to study the properties of our model. Finally, we design an interactive experiment, with users instructing the car in real-time with randomly timed and misleading instructions.

To summarize, our core contributions are:
(1) an end-to-end policy controlling a self-driving car from language and images;
(2) a hierarchical architecture reasoning about both short and long time horizons as well as both high-level inputs and low-level continuous states and actions;
(3) an implementation of interactive language-grounded driving robust to misleading user instructions.

\section{Related work}
\label{sec:related-work}
\begin{figure}[bt]
    \centering
    \includegraphics[width=0.95\linewidth]{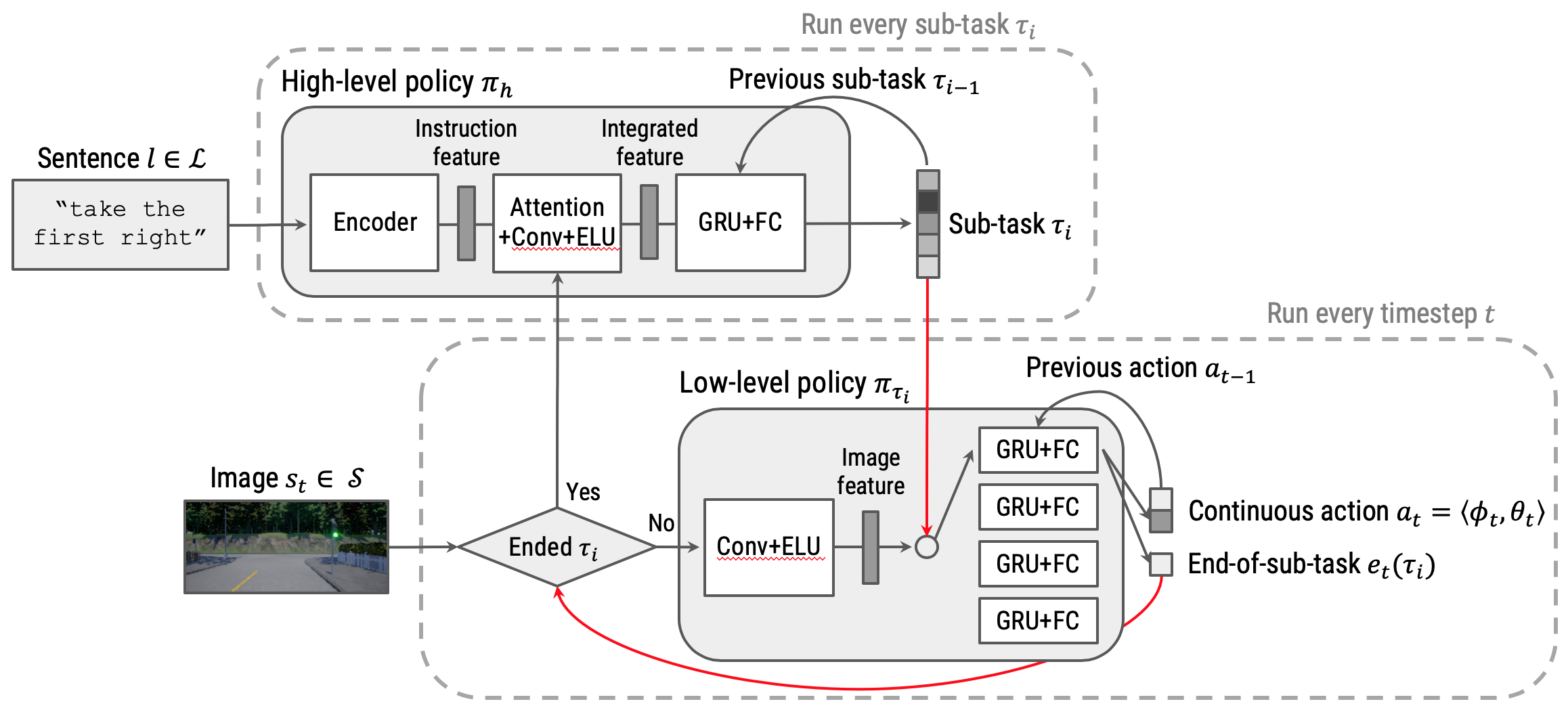}
    \caption{The proposed model for language-grounded driving. The model takes an image from the dashboard-mounted camera and a natural language instruction and generates steering and throttle values for control. Gray and red arrows represent flows of tensors and control switching signals, respectively.}
    \label{fig:entire-model-architecture}
\end{figure}

Our approach is closely related to work on Visual Question Answering (VQA), a growing area of research in which an agent is trained to move about in a home environment and find the answers to specific questions~\cite{das2018neural,chung2016iros,ma2019self,ke2019tactical,shah2018follownet,anderson2018vision}. In particular \citet{das2018neural}~proposed Neural Modular Control (NMC), which used a multi-level model to predict a ``program'' of actions that need to be taken.
While closely related, our method uses continuous state and action spaces with a realistic car model, while the vision-language navigation problem is mainly focused on dealing with complicated language expressions with relatively limited discrete state and action spaces.

Other recent work has explored learning driving policies from images. \citet{liang2018cirl} use a mixture of imitation learning and reinforcement learning via DDPG~\cite{lillicrap2015continuous}. 
\citet{codevilla2018end} proposed a system for conditional end-to-end driving.
\citet{muller2018driving} extends this work, adding a segmentation model which improves generalization performance, but largely keeping the same structure from~\citet{codevilla2018end}. \citet{paxton2017combining} also learn hierarchical policies for driving through intersections, but focus on interacting with other vehicles and do not use images.

Some work has also looked at learning representations based on images and language, but not for driving.
\citet{chaplot2018gated} proposed a Gated Attention model for learning navigation policies based on images and language, an approach we borrow for our high-level model. \citet{paxton2019prospection} learned to generate task plans and execute pick-and-place tasks. \citet{blukis2018following} learn a semantic map for navigation and demonstrate on a simulated quadrotor, which can be used to follow natural language instructions.

While we provide a manual decomposition of the task when training models as seen in some previous work~\cite{paxton2019prospection,das2018neural}, some prior work weakens these assumptions. \citet{shiarlis2018taco} propose TACO, which learns to break tasks up based only on a policy sketch. \citet{krishnan2017ddco} also propose a method for discovery of continuous actions from demonstrations, which could be applied to our problem in the future.
\citet{andreas2017modular} uses policy sketches together with curriculum reinforcement learning.

\section{Hierarchical Language-Grounded Driving Model}
\label{sec:approach}
The goal of our agent is to drive safely, following given language directions and a stream of images from a single camera.
Driving requires a very long time horizon, with high-frequency controls but low-frequency decisions.
This makes it difficult to directly apply a sequence-to-sequence approach to the problem.
Instead, we introduce a hierarchical driving model where a high-level policy $\pi_h$ chooses a series of sub-tasks $\{\tau_0,\dots,\tau_{N-1}\}$ to achieve a specified task, and low-level policies $\pi_{\tau_i}$ generate the controls necessary to achieve each sub-task $\tau_i$ in sequence.
This decomposes the problem into tractable sub-problems and enables efficient use of short data sequences for training complex, language-conditioned control policies.
This approach is similar to that taken in the vision-language navigation problem~\cite{das2018neural}, but with increased complexity of low-level controls.

Algorithm \ref{alg:execution} shows the pseudo code for execution of our language-grounded driving model. Figure \ref{fig:entire-model-architecture} shows the architecture of our model.
Consider the world $W: \mathcal{S} \times \mathcal{A} \rightarrow \mathcal{S}$ with a continuous state observation $s \in \mathcal{S}$ and a continuous action $a = \langle \phi, \theta \rangle \in \mathcal{A}$, where $\phi \in [0, 1]$ is the normalized throttle control and $\theta \in [-1, 1]$ is the normalized steering angle for the vehicle.
We assume that our state observation $s_t$ consists of an image from a dashboard-mounted camera at time $t$. Then the directions for the language-grounded driving are specified by a natural-language input $l \in \mathcal{L}$, such as ``take the next right'' or ``go straight through this intersection, then turn left.'' Finally, our problem is defined by learning a policy $\pi: \mathcal{S} \times \mathcal{L} \rightarrow \mathcal{A}$.

We break up the original problem into a two-level hierarchy by introducing a sub-task $\tau \in \mathcal{T}$ where the set of possible sub-tasks $\mathcal{T} = \{\texttt{left}, \texttt{right}, \texttt{straight}, \texttt{lanefollow}\}$. The sub-task \texttt{straight} represents the case of going straight through an intersection while $\texttt{lanefollow}$ corresponds to a policy ensuring safe lane following outside intersections. We extend $\mathcal{T}$ to include a \texttt{finish} token, indicating the end of the entire task: $\hat{\mathcal{T}} = \mathcal{T} \cup \{\texttt{finish}\}$.

With the hierarchical model, our problem is to learn a high level policy $\pi_{h}: \mathcal{S} \times \mathcal{L} \rightarrow \hat{\mathcal{T}}$, and a corresponding low-level policy $\pi_{\tau}: \mathcal{S} \times \mathcal{T} \rightarrow \mathcal{A}$.
Also, our model detects when the sub-task is achieved. This determines whether the high or the low-level policy takes control. We define the end-of-sub-task signal as $e_t (\tau_i) \in \mathcal{E} = \{\texttt{True}, \texttt{False}\}$ which is an indicator that the current sub-task $\tau_i$ is finished and the high-level policy should regain control to generate the next sub-task. Therefore, the revised low-level policy can be expressed as $\pi_{\tau}: \mathcal{S} \times \mathcal{T} \rightarrow \mathcal{A} \times \mathcal{E}$. 

\subsection{High-level policy}
\label{sssec:high-level-policy}
The high-level policy consists of an encoder, the Gated Attention (GA) unit~\cite{chaplot2018gated}, and a recurrent unit.
First, we generate a list of 50-dimensional GloVe~\cite{pennington2014glove} embedding vectors from words in the language instruction.
Then the embedding vectors are fed to a single-layer GRU~\cite{chung2014empirical} and the attention mechanism introduced in~\cite{luong2015effective} combines hidden states from the GRU to generate a single instruction feature vector.
The image is fed to a series of convolution and ELU~\cite{clevert2015fast} layers to generate an image feature vector.

Then the GA takes the instruction and image feature vectors and generates an integrated feature vector.
The GA computes attention weights from the instruction to focus on the essential part of the image feature vector.
Finally, the integrated feature vector is concatenated with another feature vector from the previous sub-task vector and fed into another single-layer GRU and a fully connected layer. We use a softmax function to generate 5-dimensional sub-task probability distribution for $\hat{\mathcal{T}}$.

\subsection{Low-level policy}
\label{sssec:low-level-policy}
Once the sub-task $\tau_i$ is determined, the low-level policy takes control from the high-level policy and generates actions required to achieve the sub-task.
First, it converts the input image to an image feature vector by applying a series of convolutional and ELU layers.
Then, sub-task probabilities determined by the high-level policy are used to select one of a few sub-task-specific GRU layers.
The activated GRU layer combined with an FC layer generates the final 2-dimensional control vector $a_t$ and the end-of-sub-task signal $e_t(\tau_i)$.

The low-level policy remains in control until the end of the sub-task indicated by the $e_t(\tau_i)$ value.
In order to make this transition robust to noisy predictions, we require that at least two out of the three recent predictions of $e_t(\tau_i)$ indicate the end of sub-task.
In our implementation, we rely on two different low-level policies for predicting $a_t$ and $e_t(\tau_i)$.

\vskip 0.2cm
\begin{minipage}[bt]{.50\textwidth}
    \tfsz
    \fbox{\parbox[c]{\hsize}{
    \label{alg:execution}
    \begin{algorithmic} 
    \REQUIRE Initial state $s_0 \in \mathcal{S}$
    \REQUIRE Language direction $l \in \mathcal{L}$
    \REQUIRE World $W: \mathcal{S} \times \mathcal{A} \rightarrow \mathcal{S}$
    \REQUIRE High-level policy $\pi_{h}: \mathcal{S} \times \mathcal{L} \rightarrow \hat{\mathcal{T}}$
    \REQUIRE Low-level policies $\pi_{\tau}: \mathcal{S} \times \mathcal{T} \rightarrow \mathcal{A} \times \mathcal{E}$
    \STATE $i, t \leftarrow 0, 0$
    \STATE $\tau_0 \leftarrow \pi_{h}(s_0, l)$
    \STATE $e_0(\tau_0) \leftarrow \texttt{False}$
    \WHILE{$\tau_i \neq \mathtt{finish}$}
        \WHILE{$e_t(\tau_i)$ == \texttt{False}}
            \STATE $a_t, e_t(\tau_i) \leftarrow \pi_{\tau_i}(s_t)$
            \STATE $s_{t+1} \leftarrow W(s_t, a_t)$
            \STATE $t \leftarrow t+1$
        \ENDWHILE
        \STATE $i \leftarrow i+1$
        \STATE $\tau_i \leftarrow \pi_{h}(s_t, l)$
    \ENDWHILE
    \RETURN{}
    \end{algorithmic}
    \captionsetup{type=algorithm}\caption{Hierarchical policies for language-grounded conditional driving}
    }}
\end{minipage}%
\hskip 0.5cm
\begin{minipage}[bt]{.45\textwidth}
\centering
\begin{tabular}{p{1.2cm} c}
    \centering \ir & \noindent\parbox[c]{0.9\hsize}{\includegraphics[width=0.7\textwidth]{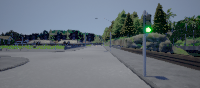} \vskip 0.1cm} \\
    \centering \ip & \noindent\parbox[c]{0.9\hsize}{\includegraphics[width=0.7\textwidth]{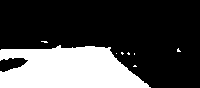} \vskip 0.1cm} \\
    \centering \ig & \noindent\parbox[c]{0.9\hsize}{\includegraphics[width=0.7\textwidth]{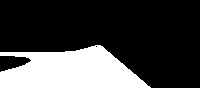} \vskip 0.1cm}
\end{tabular}
\captionsetup{type=figure}\caption{Examples of input dashboard images used in experiments: we compare performance on raw color images~(\ir) with images trained on predicted~(\ip) or ground-truth segmentation~(\ig) from  CARLA~\cite{Dosovitskiy17}.}
\label{fig:input-examples}
\end{minipage}

\section{Training and Environment}
\label{sec:environment}
We used CARLA~\cite{Dosovitskiy17} to generate data for training and run experiments. CARLA provides road annotations and an auto-pilot function.
We relied on the auto-pilot during training.

In the environments provided in CARLA, all the roads are annotated and an auto-pilot function is implemented. 
First, we deployed a roaming agent, which randomly decided a direction at each intersection, and recorded observations from the agent at 10 Hz. 
We use two towns provided by the simulator, Town1 and Town2, as in previous work~\cite{codevilla2018end,muller2018driving}.
Second, we partitioned the trajectory into a set of trajectory snippets corresponding to different sub-tasks.
State observation, action, sub-task and end-of-sub-task values, $\langle s_t, a_t, \tau_t, e_t \rangle$, were generated for the snippets. Finally, we combined language data gathered from human users with the information about the snippets to generate realistic natural language instructions corresponding to our environment. 
In following subsections, we give details of the environment, data generation, and training.

\subsection{Language generation}
We collected language data by designing a two-player driving game with human subjects.
In the game, one player was tasked with navigating the vehicle to a goal without any knowledge about the map of the world. The second player was tasked with instructing the first player about directions to goal using only natural languages.
From this data, we designed templates to generate language instructions for training the high-level policy that matched our test environments. For more detailed procedure of the language generation, please see Appendix~\ref{ssec:language}.

We generated instructions of varying complexity that would be typical for interactions between a human and an autonomous vehicle.
We grouped them into three categories (see Table~\ref{tab:sentences} for examples).
The first category (\ls) contains instructions that tell the car how the behave at the next intersection.
The second category (\ld) contains instructions about the behavior at the two upcoming intersections.
Finally, the third category (\lo) contains instructions including ordinal expressions relating to any of two upcoming intersections.

\begin{table}[bt]
    \tfsz
    \centering
    \begin{tabular}{p{0.18\columnwidth} p{0.35\columnwidth} p{0.35\columnwidth}}
    \toprule
        \ls & \ld & \lo \\
        \midrule[0.8pt]
        turn left & turn left at first and then right & you re going to take your second left up here \\
        \midrule[0.4pt]
        make a left turn & take a left here and then you re going to take a another right turn & you re going to go a little bit further for one block and make a left at the intersection \\
        \midrule[0.4pt]
        left & take your left here and then it will be a right turn ahead & take the second left \\
    \bottomrule
    \end{tabular}
    \vskip 0.1cm
    \caption{Examples of generated sentences for left turns based on data gathered from realistic interactions. Examples are grouped into three categories, depending on time horizon and complexity of the instruction.}
    \label{tab:sentences}
\end{table}

\subsection{Training and Trajectory Generation}
We collected expert trajectories in simulation and trained each model with supervision analogous to prior work~\cite{muller2018driving,das2018neural}.
First, we collected an expert trajectory, $d = \{p_t: t \in T\}$, by releasing a randomly roaming expert with a global planner and PID controller similar to~\cite{muller2018driving}, and obtained training sub-task labels based on the annotated road structure, where $p_t = \langle s_t, a_t, \tau_t \rangle$, $s_t \in \mathcal{S} = \mathbb{R}^{3 \times 200 \times 88}$ is an image used as a state observation, $a_t \in \mathcal{A}$ is an action, and $\tau_t \in \mathcal{T}$ is a sub-task label at time $t$. 

Then, we partitioned the trajectories into a set of trajectory snippets $D = \{d_i\}$ for training both low-level and high-level policies based on the sub-task labels where $d_i = \left\{ p_t: t \in R_i \right\}$ and $R_i = \left[p_i, q_i \right]$ such that $p_i \leq q_i \land p_i \in T \land q_i \in T \land \tau_a = \tau_b~ \forall{a,b} \in \left[p_i, q_i \right]$. Snippets around intersections were segmented according to which turn was taken into the \texttt{left}, \texttt{right}, and \texttt{straight} policies, corresponding to each of the three possible choices. Intersections were connected by the sub-task \texttt{lanefollow}.

\begin{figure}[bt]
\centering
\includegraphics[width=0.55\columnwidth]{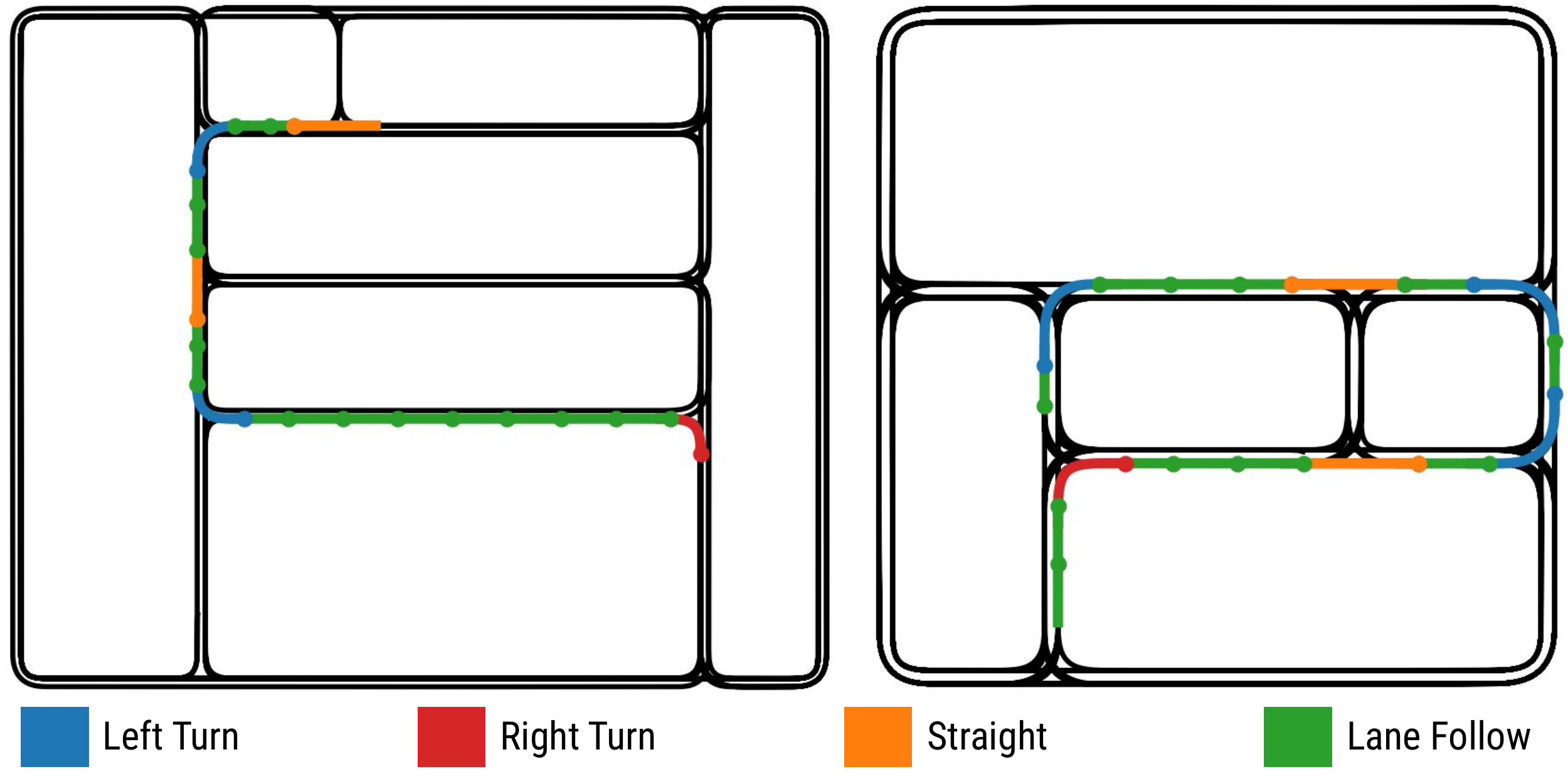}
\caption{Top-down view showing trajectory segments with sub-task annotations.}  
\label{fig:high-level-examples}
\end{figure}

In order to make the low-level policy robust, we added a margin before and after each snippet, so that each low-level policy also learns to follow the lane. We used $L^1$ loss for training the control model and binary cross entropy loss for training the end-of-sub-task model.

For high-level policies, we used three or five snippets as a longer segment which included one or two intersections and neighboring \texttt{lanefollow} snippets. Within the segment, we drew data points from boundary regions between sub-tasks for training. We use cross entropy loss for training the high-level model. Fig.~\ref{fig:high-level-examples} shows a couple of examples of road segments with sub-tasks annotations.

In addition to one front-facing camera, we put two additional cameras rotated about 14 degrees to the left and to the right, to simulate the images from drifted states.
We used three types of images to examine the effects of input modality on generalization from one town to the next, shown in see Fig.~\ref{fig:input-examples}. These are:
(1) \ir: color images from the dash-mounted camera,
(2) \ig: ground-truth binary road segmentation images from the simulator; and
(3) \ip: predicted binary road segmentation images from DeepLabv3+~\cite{deeplabv3plus2018} with Mobilenetv2~\cite{mobilenetv22018} pretrained with COCO~\cite{lin2014microsoft} and fine-tuned on the Cityscapes dataset~\cite{cordts2016cityscapes}.

\section{Experiments and Results}
\label{sec:experiments}
We perform a comprehensive evaluation of different properties of our model, and compare it to established baselines.
We begin with a quantitative evaluation of generalization abilities of the model for different types of observations.
We follow with ablation experiments and comparisons to previously published baselines.
Then, we demonstrate robustness to misleading instructions and randomly timed commands.
Finally, we show how our model can be used in interactive, real-time scenarios.
In the following evaluations, we trained the model on Town1 and tested on both Town1 and Town2.
In the training procedure, we draw fixed-length trajectories from the trajectory snippets.

\subsection{Input comparison}
\label{ssec:input-comparison}
One challenge with training policies on unstructured input such as images and language is transferring models to new environments. We explored the effects of different input modalities on performance and generalization inspired by the previous work~\cite{muller2018driving} which has shown that segmentation-based policies transfer well between environments. The results can be seen in Table~\ref{tab:ablation-hierarchical}.

\begin{table}
    \tfsz
    \centering
    \begin{tabular}{r c c c c c c}
    \toprule
    \multicolumn{1}{c}{\multirow{1}{*}{\textbf{Input Modality $\rightarrow$}}}
    &
    \multicolumn{2}{c}{\ir} &
    \multicolumn{2}{c}{\ig} & 
    \multicolumn{2}{c}{\ip}\\
    \multicolumn{1}{c}{\multirow{1}{*}{\textbf{Language Type $\downarrow$}}}
    & train & test 
    & train & test
    & train & test \\
    \midrule[0.8pt]
    \texttt{single}  & 1.000 & 1.000 & 1.000 & 1.000 & 0.982 & 1.000 \\
    \texttt{double}  & 0.809 & 0.439 & 1.000 & 0.986 & 0.976 & 0.874 \\
    \texttt{ordinal} & 0.813 & 0.333 & 1.000 & 0.938 & 1.000 & 0.938 \\
    \texttt{all}     & 0.880 & 0.613 & 1.000 & 0.970 & 0.982 & 0.926 \\
    \bottomrule
    \end{tabular}
    \vskip0.2cm
    \caption{Comparison of results for three different input modalities: ground-truth segmentation \ig, predicted segmentation \ip, and raw color images \ir.}
    \label{tab:ablation-hierarchical}
\end{table}

We performed language-grounded driving by starting from the beginning of the trajectory, given a randomly sampled sentence, and measure the rate of successful episodes.
Overall, the model achieved almost perfect results for all levels of language complexity as long as ground truth segmentation was used as observations, even when generalizing across different environments.
The performance dropped slightly, when predicted segmentation was used.
Finally, raw color images resulted in largest performance drop when the model was transferred across environments, despite good performance on the training environment.
Table \ref{tab:ablation-hierarchical} shows the quantitative evaluation result. The average performance drop from Town1 to Town2 with~\ir~is about $30.13\%$ while the averages performance drops of~\ip~and~\ig~are about $5.038\%$ and $1.301\%$.
This reaffirms good generalization performance of semantic segmentation.
This trend was primarily noticeable for language instructions of highest complexity (\lo).
A full comparison with model ablations and input modalities is provided in Table~\ref{tab:ablation-hierarchical-appendix} in Appendix.

\subsection{Model Comparison}
\label{ssec:model-comparison}

\begin{table}
    \tfsz
    \centering
    \begin{tabular}{l c c c c c c c c}
    \toprule
    \multicolumn{1}{c}{\multirow{1}{*}{\textbf{Language Type} $\rightarrow$}} &
    \multicolumn{2}{c}{\texttt{single}} &
    \multicolumn{2}{c}{\texttt{double}} &
    \multicolumn{2}{c}{\texttt{ordinal}} &
    \multicolumn{2}{c}{\texttt{all}} \\
    \multicolumn{1}{c}{\multirow{1}{*}{\textbf{Model} $\downarrow$}}
    & train & test & train & test & train & test & train & test\\
    \midrule[0.8pt]
    Single policy 
    & 0.721 & 0.583 & 0.033 & 0.000 & 0.028 & 0.000 & 0.288 & 0.197 \\
    Single policy with history
    & 0.684 & 0.667 & 0.077 & 0.000 & 0.142 & 0.000 & 0.311 & 0.225 \\
    \midrule[0.4pt]
    NMC~\cite{das2018neural} 
    & 0.218 & 0.208 & 0.000 & 0.000 & 0.000 & 0.000 & 0.081 & 0.070 \\
    \midrule[0.4pt]
    $\mb$: hierarchical baseline 
    & \textbf{1.000} & \textbf{1.000} & \textbf{1.000} & \textbf{1.000} & 0.969 & 0.906 & 0.996 & 0.986 \\
    $\mi$: $\mb$ with images 
    & \textbf{1.000} & \textbf{1.000} & 0.979 & 0.982 & \textbf{1.000} & 0.813 & 0.990 & \textbf{0.991} \\
    $\mih$: $\mi$ with history \textbf{(full model)} 
    & \textbf{1.000} & \textbf{1.000} & \textbf{1.000} & 0.960 & \textbf{1.000} & \textbf{0.938} & \textbf{1.000} & 0.970 \\
    $\mihg$: $\mih$ with gated attention 
    & 0.984 & \textbf{1.000} & 0.976 & 0.943 & 0.972 & 0.886 & 0.979 & 0.954 \\
    \bottomrule
    \end{tabular}
    \vskip0.2cm
    \caption{Comparison of our method to both a single policy and a Neural Modular Control (NMC)~\cite{das2018neural} baseline, and ablation of several different key components, given ground-truth road segmentation as input (\texttt{gs}.) Models used in ablation, $\mb$, $\mi$, and $\mih$, are explained in Section~\ref{ssec:model-comparison}. $\mihg$ replaces the original low-level model with Gated Attention model~\cite{chaplot2018gated}.}
    \label{tab:comparison}
\end{table}

We compared our model ($\mih$) with three variants of the model and two baseline models, a single policy and a Neural Modular Control (NMC)~\cite{das2018neural}, given ground-truth road segmentation as input (\ig).
A single policy was implemented by extending a high-level policy to directly predict actions.
We implemented NMC without attention for post-navigation question answering.
The first variant ($\mb$) is a hierarchical baseline model with a high-level policy which only takes language instruction as its input.
A high-level policy in the second variant ($\mi$) takes the image along with the language instruction but it does not use the sub-task history.
Our model ($\mih$) uses the language instruction, the image, and the sub-task history in the high-level policy.
The last variant ($\mihg$) uses the same high-level model as our model but it replaces the original low-level policy using a few sub-task-specific GRU layers by a new low-level policy which is conditioned by a sub-task label using GA.
A single policy was implemented by extending a high-level policy to directly predict actions.
Table~\ref{tab:comparison} shows that all variants of our model outperformed the baselines.

We see that both baselines struggled to interpret more complex commands (\ld~or~\lo).
The single policy could learn a turning behavior from fixed-length sub-trajectories but failed to learn to distinguish multiple turns and plan a series of turns.
This is understandable, given the length of the trajectories (hundreds of frames), which do not fit in a single recurrent unit.
The hierarchical decomposition of the task in our model reduces the complexity of the problem and makes the model trainable with long-time horizon data.
The NMC baseline performed poorly even for simple directions (\ls).
Lack of an attention mechanisms resulted in poor performance of end-of-sub-task and sub-task prediction.

Among our model and two hierarchical variants of the models, $\mih$, $\mb$ and $\mi$, we could not see a huge performance gap.
Transition between sub-tasks is highly dependent on the end-of-sub-task value from low-level policy and that gives the model with a simple high-level policy high performance.

In addition, we replace the original low-level model with the gated-attention model, $\mihg$, which takes sub-task values as a conditional input.
Though this conditional low-level model performs slightly worse than the original model,  $\mih$, its performance is comparable to other ablation models.
It implies that switching between a fixed number of special layers is not necessarily needed; for higher-level tasks in future, we can generalize the intermediate representation not restricted to a fixed number of sub-tasks.
\subsection{Misleading Instructions}

\begin{table}
    \tfsz
    \centering
    \begin{tabular}{l c c c c c c}
    \toprule
    \multicolumn{1}{c}{\multirow{1}{*}{\textbf{Language Type} $\rightarrow$}} & 
    \multicolumn{2}{c}{\texttt{single}} &
    \multicolumn{2}{c}{\texttt{double}} &
    \multicolumn{2}{c}{\texttt{all}} \\
    \multicolumn{1}{c}{\multirow{1}{*}{\textbf{Model} $\downarrow$}} & 
    train & test & train & test & train & test \\
    \midrule[0.8pt]
    $\mb$: hierarchical baseline & 
    \textbf{1.000} & \textbf{1.000} & 0.758 & 0.652 & 0.828 & 0.742 \\
    $\mi$: $\mb$ with images & 
    \textbf{1.000} & \textbf{1.000} & \textbf{1.000} & \textbf{0.957} & \textbf{1.000} & \textbf{0.968} \\
    $\mih$: $\mi$ with sub-task history \textbf{(full model)} & 
    \textbf{1.000} & \textbf{1.000} & \textbf{1.000} & 0.928 & \textbf{1.000} & 0.946 \\
    \bottomrule
    \end{tabular}
    \vskip0.2cm
    \caption{Evaluation of our approach for misleading language instructions (e.g. ``go straight'' when no straight road exists). We used ground-truth segmentation images \ig~as input.}
    \label{tab:ablation-misleading-instructions}
\end{table}

Realistic natural language instructions are often inconsistent and ambiguous.
For an autonomous system, it is paramount to handle such instructions and generate only safe behavior.
To evaluate our model in such conditions, we generated misleading language instructions containing directions not possible to execute given the current world map (e.g. ``go straight" for a T-shaped intersection).
In such cases, we expect a safe behavior and choose to train the model to stop for impossible straight directions or go straight for impossible turns.
We evaluated the resulting model only on misleading instructions.

Table \ref{tab:ablation-misleading-instructions} shows the quantitative evaluation of our model for misleading language directions using ground-truth segmentation images as input.
Usage of the image in the high-level policy seemed to be an important factor on performance.
Performance of the model $\mb$ on complicated language directions is degraded in comparison to other models which use images in the high-level policy.
This demonstrates the importance of using image in decision making when it has to deal with misleading language instructions.
Our model shows the robustness to misleading instructions.

\begin{figure}[bt]
\centering
\includegraphics[width=0.85\columnwidth]{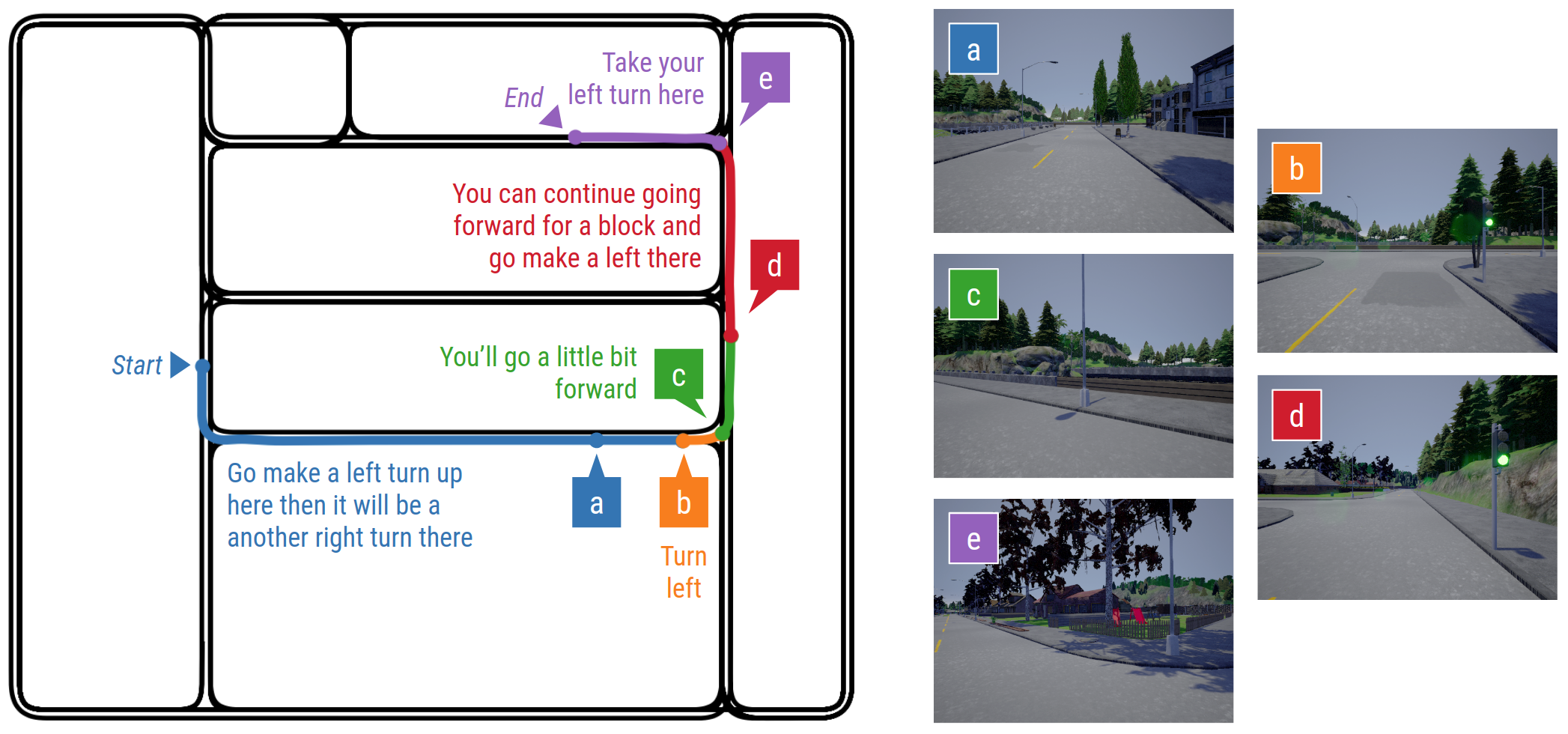}
\caption{An example of interactive driving. The trajectory and the instructions provided by the user are shown on the left. The right side shows images corresponding to the indicated points along the trajectory.}
\label{fig:interactive}
\end{figure}

\subsection{Interactive Driving}
In realistic settings, a self-driving car will receive instructions from the user at different moments in time, even if the car is currently executing a previous command.
To illustrate the robustness of our model to imperfectly timed commands as well as random interruptions, we designed an interactive, real-time driving protocol, where users could provide natural language instructions at any moment in time.
The agent interrupts the current plan whenever a new instruction s received.
Here, we present and analyse an example of such experiment (see Fig.~\ref{fig:interactive}).

The experiment began with the instruction ``Go make a left turn up here then it will be a another right turn there."
Then the user interrupted the execution of the commands three times at random moments, often while a sub-task such as left turn is currently being executed.
This interactive driving example shows that our agent can be used to drive continuously according to user directions, even when frequently interrupted or when given inconsistent commands.
Thanks to its hierarchical structure, our model is less sensitive to timing issues; it will complete the current maneuver before executing the next command.

\section{Conclusion}
\label{sec:conclusion}
We showed a system for linguistic control of a self-driving vehicle from images, and provide an ablation analysis of which components of the network are important for providing the best performance including generalization to new environments. In particular, we showed our model improves on related prior work for visual question answering~\cite{das2018neural} and extends work in driving using conditional policies. Our future work will focus on even more complex language expressions, with emphasis on objects in the environment.


\acknowledgments{This work was supported by the UW Reality Lab, Facebook, Google, and Huawei. We would also like to thank NVIDIA for generously providing DGX used for this research via the NVIDIA Robotics Lab and the UW NVIDIA AI Lab (NVAIL) and Samsung Scholarship.}

\bibliography{papers}  

\begin{thebibliography}{28}
\providecommand{\natexlab}[1]{#1}
\providecommand{\url}[1]{\texttt{#1}}
\expandafter\ifx\csname urlstyle\endcsname\relax
  \providecommand{\doi}[1]{doi: #1}\else
  \providecommand{\doi}{doi: \begingroup \urlstyle{rm}\Url}\fi

\bibitem[Bojarski et~al.(2016)Bojarski, Del~Testa, Dworakowski, Firner, Flepp,
  Goyal, Jackel, Monfort, Muller, Zhang, et~al.]{bojarski2016end}
M.~Bojarski, D.~Del~Testa, D.~Dworakowski, B.~Firner, B.~Flepp, P.~Goyal, L.~D.
  Jackel, M.~Monfort, U.~Muller, J.~Zhang, et~al.
\newblock End to end learning for self-driving cars.
\newblock \emph{arXiv preprint arXiv:1604.07316}, 2016.

\bibitem[Xu et~al.(2017)Xu, Gao, Yu, and Darrell]{xu2017end}
H.~Xu, Y.~Gao, F.~Yu, and T.~Darrell.
\newblock End-to-end learning of driving models from large-scale video
  datasets.
\newblock In \emph{Proceedings of the IEEE conference on computer vision and
  pattern recognition}, pages 2174--2182, 2017.

\bibitem[Codevilla et~al.(2018)Codevilla, Miiller, L{\'o}pez, Koltun, and
  Dosovitskiy]{codevilla2018end}
F.~Codevilla, M.~Miiller, A.~L{\'o}pez, V.~Koltun, and A.~Dosovitskiy.
\newblock End-to-end driving via conditional imitation learning.
\newblock In \emph{2018 IEEE International Conference on Robotics and
  Automation (ICRA)}, pages 1--9. IEEE, 2018.

\bibitem[M{\"u}ller et~al.(2018)M{\"u}ller, Dosovitskiy, Ghanem, and
  Koltun]{muller2018driving}
M.~M{\"u}ller, A.~Dosovitskiy, B.~Ghanem, and V.~Koltun.
\newblock Driving policy transfer via modularity and abstraction.
\newblock \emph{arXiv preprint arXiv:1804.09364}, 2018.

\bibitem[Paxton et~al.(2019)Paxton, Bisk, Thomason, Byravan, and
  Fox]{paxton2019prospection}
C.~Paxton, Y.~Bisk, J.~Thomason, A.~Byravan, and D.~Fox.
\newblock Prospection: Interpretable plans from language by predicting the
  future.
\newblock \emph{IEEE International Conference on Robotics and Automation
  (ICRA)}, 2019.

\bibitem[Dosovitskiy et~al.(2017)Dosovitskiy, Ros, Codevilla, Lopez, and
  Koltun]{Dosovitskiy17}
A.~Dosovitskiy, G.~Ros, F.~Codevilla, A.~Lopez, and V.~Koltun.
\newblock {CARLA}: {An} open urban driving simulator.
\newblock In \emph{Proceedings of the 1st Annual Conference on Robot Learning},
  pages 1--16, 2017.

\bibitem[Das et~al.(2018)Das, Gkioxari, Lee, Parikh, and Batra]{das2018neural}
A.~Das, G.~Gkioxari, S.~Lee, D.~Parikh, and D.~Batra.
\newblock Neural modular control for embodied question answering.
\newblock \emph{arXiv preprint arXiv:1810.11181}, 2018.

\bibitem[Chung* et~al.(2016)Chung*, Pronobis*, Cakmak, Fox, and
  Rao]{chung2016iros}
M.~J.-Y. Chung*, A.~Pronobis*, M.~Cakmak, D.~Fox, and R.~P.~N. Rao.
\newblock Autonomous question answering with mobile robots in human-populated
  environments.
\newblock In \emph{Proceedings of the 2016 IEEE/RSJ International Conference on
  Intelligent Robots and Systems (IROS)}, 2016.

\bibitem[Ma et~al.(2019)Ma, Lu, Wu, AlRegib, Kira, Socher, and
  Xiong]{ma2019self}
C.-Y. Ma, J.~Lu, Z.~Wu, G.~AlRegib, Z.~Kira, R.~Socher, and C.~Xiong.
\newblock Self-monitoring navigation agent via auxiliary progress estimation.
\newblock \emph{arXiv preprint arXiv:1901.03035}, 2019.

\bibitem[Ke et~al.(2019)Ke, Li, Bisk, Holtzman, Gan, Liu, Gao, Choi, and
  Srinivasa]{ke2019tactical}
L.~Ke, X.~Li, Y.~Bisk, A.~Holtzman, Z.~Gan, J.~Liu, J.~Gao, Y.~Choi, and
  S.~Srinivasa.
\newblock Tactical rewind: Self-correction via backtracking in
  vision-and-language navigation.
\newblock \emph{arXiv preprint arXiv:1903.02547}, 2019.

\bibitem[Shah et~al.(2018)Shah, Fiser, Faust, Kew, and
  Hakkani-Tur]{shah2018follownet}
P.~Shah, M.~Fiser, A.~Faust, J.~C. Kew, and D.~Hakkani-Tur.
\newblock Follownet: Robot navigation by following natural language directions
  with deep reinforcement learning.
\newblock \emph{arXiv preprint arXiv:1805.06150}, 2018.

\bibitem[Anderson et~al.(2018)Anderson, Wu, Teney, Bruce, Johnson,
  S{\"u}nderhauf, Reid, Gould, and van~den Hengel]{anderson2018vision}
P.~Anderson, Q.~Wu, D.~Teney, J.~Bruce, M.~Johnson, N.~S{\"u}nderhauf, I.~Reid,
  S.~Gould, and A.~van~den Hengel.
\newblock Vision-and-language navigation: Interpreting visually-grounded
  navigation instructions in real environments.
\newblock In \emph{Proceedings of the IEEE Conference on Computer Vision and
  Pattern Recognition}, pages 3674--3683, 2018.

\bibitem[Liang et~al.(2018)Liang, Wang, Yang, and Xing]{liang2018cirl}
X.~Liang, T.~Wang, L.~Yang, and E.~Xing.
\newblock {CIRL}: Controllable imitative reinforcement learning for
  vision-based self-driving.
\newblock In \emph{Proceedings of the European Conference on Computer Vision
  (ECCV)}, pages 584--599, 2018.

\bibitem[Lillicrap et~al.(2015)Lillicrap, Hunt, Pritzel, Heess, Erez, Tassa,
  Silver, and Wierstra]{lillicrap2015continuous}
T.~P. Lillicrap, J.~J. Hunt, A.~Pritzel, N.~Heess, T.~Erez, Y.~Tassa,
  D.~Silver, and D.~Wierstra.
\newblock Continuous control with deep reinforcement learning.
\newblock \emph{arXiv preprint arXiv:1509.02971}, 2015.

\bibitem[Paxton et~al.(2017)Paxton, Raman, Hager, and
  Kobilarov]{paxton2017combining}
C.~Paxton, V.~Raman, G.~D. Hager, and M.~Kobilarov.
\newblock Combining neural networks and tree search for task and motion
  planning in challenging environments.
\newblock \emph{Intelligent Robots and Systems (IROS), 2017 IEEE/RSJ
  International Conference on}, 2017.

\bibitem[Chaplot et~al.(2018)Chaplot, Sathyendra, Pasumarthi, Rajagopal, and
  Salakhutdinov]{chaplot2018gated}
D.~S. Chaplot, K.~M. Sathyendra, R.~K. Pasumarthi, D.~Rajagopal, and
  R.~Salakhutdinov.
\newblock Gated-attention architectures for task-oriented language grounding.
\newblock In \emph{Thirty-Second AAAI Conference on Artificial Intelligence},
  2018.

\bibitem[Blukis et~al.(2018)Blukis, Brukhim, Bennett, Knepper, and
  Artzi]{blukis2018following}
V.~Blukis, N.~Brukhim, A.~Bennett, R.~A. Knepper, and Y.~Artzi.
\newblock Following high-level navigation instructions on a simulated
  quadcopter with imitation learning.
\newblock \emph{arXiv preprint arXiv:1806.00047}, 2018.

\bibitem[Shiarlis et~al.(2018)Shiarlis, Wulfmeier, Salter, Whiteson, and
  Posner]{shiarlis2018taco}
K.~Shiarlis, M.~Wulfmeier, S.~Salter, S.~Whiteson, and I.~Posner.
\newblock {TACO}: Learning task decomposition via temporal alignment for
  control.
\newblock \emph{arXiv preprint arXiv:1803.01840}, 2018.

\bibitem[Krishnan et~al.(2017)Krishnan, Fox, Stoica, and
  Goldberg]{krishnan2017ddco}
S.~Krishnan, R.~Fox, I.~Stoica, and K.~Goldberg.
\newblock Ddco: Discovery of deep continuous options for robot learning from
  demonstrations.
\newblock \emph{arXiv preprint arXiv:1710.05421}, 2017.

\bibitem[Andreas et~al.(2017)Andreas, Klein, and Levine]{andreas2017modular}
J.~Andreas, D.~Klein, and S.~Levine.
\newblock Modular multitask reinforcement learning with policy sketches.
\newblock In \emph{Proceedings of the 34th International Conference on Machine
  Learning-Volume 70}, pages 166--175. JMLR. org, 2017.

\bibitem[Pennington et~al.(2014)Pennington, Socher, and
  Manning]{pennington2014glove}
J.~Pennington, R.~Socher, and C.~Manning.
\newblock Glo{V}e: global vectors for word representation.
\newblock In \emph{Proceedings of the 2014 conference on empirical methods in
  natural language processing (EMNLP)}, pages 1532--1543, 2014.

\bibitem[Chung et~al.(2014)Chung, Gulcehre, Cho, and
  Bengio]{chung2014empirical}
J.~Chung, C.~Gulcehre, K.~Cho, and Y.~Bengio.
\newblock Empirical evaluation of gated recurrent neural networks on sequence
  modeling.
\newblock \emph{arXiv preprint arXiv:1412.3555}, 2014.

\bibitem[Luong et~al.(2015)Luong, Pham, and Manning]{luong2015effective}
M.-T. Luong, H.~Pham, and C.~D. Manning.
\newblock Effective approaches to attention-based neural machine translation.
\newblock \emph{arXiv preprint arXiv:1508.04025}, 2015.

\bibitem[Clevert et~al.(2015)Clevert, Unterthiner, and
  Hochreiter]{clevert2015fast}
D.-A. Clevert, T.~Unterthiner, and S.~Hochreiter.
\newblock Fast and accurate deep network learning by exponential linear units
  (elus).
\newblock \emph{arXiv preprint arXiv:1511.07289}, 2015.

\bibitem[Chen et~al.(2018)Chen, Zhu, Papandreou, Schroff, and
  Adam]{deeplabv3plus2018}
L.-C. Chen, Y.~Zhu, G.~Papandreou, F.~Schroff, and H.~Adam.
\newblock Encoder-decoder with atrous separable convolution for semantic image
  segmentation.
\newblock In \emph{ECCV}, 2018.

\bibitem[Sandler et~al.(2018)Sandler, Howard, Zhu, Zhmoginov, and
  Chen]{mobilenetv22018}
M.~Sandler, A.~Howard, M.~Zhu, A.~Zhmoginov, and L.-C. Chen.
\newblock Mobile{N}et{V2}: Inverted residuals and linear bottlenecks.
\newblock In \emph{CVPR}, 2018.

\bibitem[Lin et~al.(2014)Lin, Maire, Belongie, Hays, Perona, Ramanan,
  Doll{\'a}r, and Zitnick]{lin2014microsoft}
T.-Y. Lin, M.~Maire, S.~Belongie, J.~Hays, P.~Perona, D.~Ramanan,
  P.~Doll{\'a}r, and C.~L. Zitnick.
\newblock Microsoft {COCO}: Common objects in context.
\newblock In \emph{European conference on computer vision}, pages 740--755.
  Springer, 2014.

\bibitem[Cordts et~al.(2016)Cordts, Omran, Ramos, Rehfeld, Enzweiler, Benenson,
  Franke, Roth, and Schiele]{cordts2016cityscapes}
M.~Cordts, M.~Omran, S.~Ramos, T.~Rehfeld, M.~Enzweiler, R.~Benenson,
  U.~Franke, S.~Roth, and B.~Schiele.
\newblock The {C}ityscapes dataset for semantic urban scene understanding.
\newblock In \emph{Proceedings of the IEEE conference on computer vision and
  pattern recognition}, pages 3213--3223, 2016.

\end{thebibliography}

\appendix
\section*{Appendix}
\renewcommand{\thesubsection}{\Alph{subsection}}
\setcounter{table}{0}
\renewcommand{\thetable}{A\arabic{table}}

\label{sec:appendix}
\subsection{Language generation}
\label{ssec:language}
To create the language dataset, we originally conducted a two-player game with human subjects to collect speech signals of commands and corresponding driving controls. In the game, two players are asked to collaboratively drive a car to reach three randomly spawned goals. While one player drives a car without knowing where the destination is, the other player reads a map and gives direction to the driver. After transcribing the collected audio data, we removed the sentences with actions that cannot be taken in the current environment and removed expressions mentioning objects or structures. Then we divide expressions into prefix, body, and suffix and cluster those expressions to transform the sentence into templates. Finally, we generated sentences for each combination of sub-tasks with the templates. In the implementation, we used a keyword to represent each type of combination.

We counted the number of expressions in the raw dataset for each. Table \ref{tab:sentence-count} shows the number of sentences for each keyword.
The keyword ‘\texttt{other}’ and ‘\texttt{extra}’ represents the sentences contain the actions that cannot be taken in the current environment and the sentences that do not have any meaningful commands, respectively. The total of the counted expressions is 4889 and 4600 sentences have single command {‘\texttt{left}’, ‘\texttt{right}’, ‘\texttt{straight}’, ‘\texttt{other}’, ‘\texttt{extra}’}.

\newcommand{\dd}[1]{\ifthenelse{\equal{#1}{l}}{left}{\ifthenelse{\equal{#1}{r}}{right}{\ifthenelse{\equal{#1}{s}}{straight}{\ifthenelse{\equal{#1}{1}}{first}{\ifthenelse{\equal{#1}{2}}{second}{#1}}}}}}

\newcommand{\keyword}[2][]{%
  \ifthenelse{\isempty{#1}}%
    {\texttt{\dd{#2}}} 
    {\texttt{\dd{#1}},\texttt{\dd{#2}}} 
}

\newcommand{\kc}{}
\newcommand{\kl}{\texttt{left}}
\newcommand{\kr}{\texttt{right}}
\newcommand{\ks}{\texttt{straight}}
\newcommand{\kf}{\texttt{first}}
\newcommand{\ke}{\texttt{second}}

\begin{table}
    \tfsz
    \centering
    \begin{tabular}{lcc}
    \toprule
    \multicolumn{1}{c}{\multirow{2}{*}{\textbf{Keywords}}} &
    \multicolumn{2}{c}{\textbf{Sources}} \\
    & game & templates \\
    \midrule[0.8pt]
    \keyword{l} & 1,093 & 150 \\
    \keyword{r} & 1,016 & 150 \\
    \keyword{s} & 1,199 & 454 \\
    \midrule[0.4pt]
    \keyword[l]{l} & 3 & 269,550 \\
    \keyword[l]{r} & 20 & 135,000 \\
    \keyword[l]{s} & 22 & 408,600 \\
    \keyword[r]{l} & 28 & 135,000 \\
    \keyword[r]{r} & 2 & 269,550 \\
    \keyword[r]{s} & 14 & 408,600 \\
    \keyword[s]{s} & 1 & 85 \\
    \midrule[0.4pt]
    \keyword[1]{l} &  9 & 102,150 \\
    \keyword[1]{r} &  4 & 102,150 \\
    \keyword[2]{l} & 97 & 105,450 \\
    \keyword[2]{r} & 89 & 105,450 \\
    \midrule[0.4pt]
    \keyword{other} & 913 & N/A \\
    \keyword{extra} & 379 & N/A \\
    \bottomrule
    \end{tabular}
    \vskip0.2cm
    \caption{Number of sentences collected from the preliminary two-player driving game (game) and the templates for training (templates). From the game, we also classified sentences which are out of actions defined in the environment we used in the training as \texttt{other} and sentences which do not contain meaningful commands as \texttt{extra}.}
    \label{tab:sentence-count}
\end{table}

This high percentage of single command is due to the nature of the language in the driving setting where the reactive instruction should be given within a short amount of time. This shows that concentrating on instructive sentences is a reasonable approach in the context of driving. Another point worth noting on the dataset is that people make a lot of mistakes in commanding or driving. Sometimes a commander repeats the same command until the driver finishes that action or cancels previous actions by adding a new command. Manual pruning was necessary to make the dataset feasible to train on. As a trade-off, the distribution of sentences can be made more realistic than that coming from pre-recorded driving trajectories.

As a result of this dataset imbalance, we augment natural-language phrases according to a couple of simple rules. For the sentences with two commands, we concatenated expressions from a single keyword. The dictionary shows the 14 keywords we used in the paper and the corresponding number of sentences is shown in Table \ref{tab:sentence-count}.

When we use these sentences in the training, we draw a sentence from these lists with uniform distribution. For ordinary keywords, two groups of lists were used: one from the direct combination of two sentences of single keywords and the other one from the replacement of the keyword, such as replacing ‘\texttt{left}’ with ‘\texttt{second left}’. In the training, for those with multiple groups, the group is first drawn and then the sentence is drawn from the group.

\subsection{Language examples}
\label{ssec:language-examples}
We show two examples of transcribed speech data from the preliminary two-player game experiments. Note that certain types of behavior such as going backward, reaching the target, and slowing down were excluded from the training dataset.

\begin{table}
    \begin{tabular}{p{\columnwidth}}
    \toprule 
    ``oh there's a map all right go straight", ``and you're going to turn right", ``that's good keep going straight", ``and take your first left", ``and slow down", ``all right can you see the green square", ``great", ``okay so now you want to go straight", ``and you'll take a left at the first building", ``that's good that's good keep going straight", ``and take a left", ``and take a right", ``now straight", ``and take a left", ``went a little too far so reverse and back it up", ``all right you doing good", ``go a little bit forward", ``yep there it is", ``you got it", ``okay so now you're going to want to turn around", ``you're going to back it up a little bit", ``looking good no collisions so far", ``all right now you'll take a right", ``yep", ``now go straight", ``now take a left", ``take a right", ``go straight as fast as you can", ``and you'll take a left", ``now right", ``and the exit is right up here", ``congratulations". \\
    \midrule 
     ``go straight", ``slow down a little bit", ``make a right turn", ``it's going to be a narrow street so go straight", ``and then you're going to make a left turn when you see the first", ``go straight", ``and make a left turn here", ``make a left turn", ``and go straight", ``and do you see the green spot", ``park there", ``okay", ``go straight", ``turn left turn here", ``and another left turn", ``and you're going to make a right turn here", ``and make another left turn", ``go straight", ``just go straight", ``and make another left turn", ``left turn", ``make another right turn right turn", ``go straight", ``skip this", ``and then make a left turn here left turn", ``left turn", ``left", ``and park there", ``wait for me", ``can you go back", ``reverse", ``and then left turn", ``go little more little more", ``go back back", ``back it out a little more", ``good job", ``okay go straight", ``to your left side to your left side", ``go straight", ``keep going go straight", ``pass the street intersection and then go", ``go straight", ``yeah can you go little faster", ``and then make a left turn here", ``okay try your best", ``make a left turn", ``left", ``and you're going to make another right turn right turn here right right", ``okay", ``go straight just keep going", ``pass this", ``okay slow down a little bit", ``and you going to make a left turn okay", ``go straight", ``and then make a left turn", ``left here and then left", ``make a right turn right away", ``right here right here", ``and then another right", ``right slow down slow down", ``okay go straight", ``and then the green will be on your left side left side", ``cool we are done".
    \\ 
    \bottomrule
    \end{tabular}
    \vskip0.2cm
    \caption{Language from two instances of the preliminary two-player driving game.}
    \label{tab:language-examples}
\end{table}

\begin{table}
    \tfsz
    \centering
    \begin{tabular}{l r c c c c c c}
    \toprule
    \multicolumn{1}{c}{\multirow{3}{*}{\textbf{Model}}} &
    \multicolumn{1}{c}{\multirow{3}{*}{\textbf{Language Type}}} &
    \multicolumn{6}{c}{\multirow{1}{*}{\textbf{Input Modality}}} \\
    & &
    \multicolumn{2}{c}{\ir} &
    \multicolumn{2}{c}{\ig} &
    \multicolumn{2}{c}{\ip}\\
    & & train & test & train & test & train & test\\
    \midrule[0.8pt]
    \multirow{4}{*}{$\mb$: hierarchical baseline} 
    & \texttt{single}  & 1.000 & 0.958 & 1.000 & 1.000 & 1.000 & 1.000 \\
& \texttt{double}  & 0.763 & 0.437 & 1.000 & 1.000 & 0.893 & 0.904 \\
& \texttt{ordinal} & 0.813 & 0.490 & 0.969 & 0.906 & 0.938 & 0.813 \\
& \texttt{all}     & 0.858 & \textbf{0.621} & 0.996 & 0.986 & 0.939 & 0.923 \\
\midrule[0.4pt]
\multirow{4}{*}{$\mi$: $\mb$ with image} 
& \texttt{single}  & 1.000 & 0.958 & 1.000 & 1.000 & 1.000 & 1.000 \\
& \texttt{double}  & 0.821 & 0.411 & 0.979 & 0.982 & 0.945 & 0.856 \\
& \texttt{ordinal} & 0.674 & 0.344 & 1.000 & 1.000 & 1.000 & 0.813 \\
& \texttt{all}     & 0.867 & 0.586 & 0.990 & \textbf{0.991} & 0.973 & 0.898 \\
\midrule[0.4pt]
\multirow{4}{*}{$\mih$ \textbf{(full model)}}
& \texttt{single}  & 1.000 & 1.000 & 1.000 & 1.000 & 0.982 & 1.000 \\
& \texttt{double}  & 0.809 & 0.439 & 1.000 & 0.986 & 0.976 & 0.874 \\
& \texttt{ordinal} & 0.813 & 0.333 & 1.000 & 0.938 & 1.000 & 0.938 \\
& \texttt{all}     & \textbf{0.880} & 0.613 & \textbf{1.000} & 0.970 & \textbf{0.982} & \textbf{0.926} \\
    \bottomrule
    \end{tabular}
    \vskip0.2cm
    \caption{Comparison of three different input modalities: ground-truth segmentation \ig, predicted segmentation \ip, and raw color images \ir. The highest values from \texttt{all} language type are highlighted. Models used in ablation, $\mb$, $\mi$ and $\mih$, are described in Section \ref{ssec:model-comparison}.}
    \label{tab:ablation-hierarchical-appendix}
\end{table}

\end{document}